\documentclass{article}

\usepackage{arxiv}

\usepackage[utf8]{inputenc} 
\usepackage[T1]{fontenc}    
\usepackage{hyperref}      
\usepackage{url}           
\usepackage{booktabs}      
\usepackage{amsfonts}     
\usepackage{nicefrac}      
\usepackage{microtype}     
\usepackage{subfigure,multirow}
\usepackage{tikz}
\usetikzlibrary{positioning, fit, arrows.meta, shapes}
\usepackage{filecontents,pgfplots,pgfplotstable}

\newcommand\scalemath[2]{\scalebox{#1}{\mbox{\ensuremath{\displaystyle #2}}}}
    
\title{Shuffling Recurrent Neural Networks}

\author{Michael Rotman \\
  Tel Aviv University \\
\\\And
  Lior Wolf \\
  Facebook AI Research and Tel Aviv University \\
}
\date{}
\begin{document}

\maketitle

\begin{abstract}
  We propose a novel recurrent neural network model, where the hidden state $h_t$ is obtained by permuting the vector elements of the previous hidden state $h_{t-1}$ and adding the output of a learned function $b(x_t)$ of the input $x_t$ at time $t$. In our model, the prediction is given by a second learned function, which is applied to the hidden state $s(h_t)$. The method is easy to implement, extremely efficient, and does not suffer from vanishing nor exploding gradients. In an extensive set of experiments, the method shows competitive results, in comparison to the leading literature baselines.
\end{abstract}

\section{Introduction}
\label{sec:introduction}

Recurrent Neural Networks (RNN) architectures have been successful in solving sequential or time dependent problems. Such methods maintain a latent representation, commonly referred to as the ``hidden state'', and apply the same learned functions repeatedly to the input at each time step, as well as to the current hidden state. 

A well-known challenge with RNNs, is that of exploding or vanishing gradients. The various methods that were devised in order to solve this problem can be roughly divided into two groups. The first group utilizes a gating mechanism to stabilize the gradient flow between subsequent hidden states~\cite{Hochreiter:1997:LSM:1246443.1246450,cho-etal-2014-learning}, whereas the second group focuses on preserving the norm of the hidden states by employing constraints on the family of matrices used as the network's parameters~\cite{urnn,DBLP:journals/corr/HenaffSL16}. 

An alternative view of the problem of exploding and vanishing gradient considers it as the symptom of a deeper issue and not as the root cause. Current RNN architectures perform a matrix multiplication operation, with learned weights, over previously seen hidden states during each time step. Therefore, inputs appearing in different times are processed using different powers of the weight matrices (with interleaving non-linearities):  the first input of a sequence of length $T$ is processed by the same learned sub-network $T$ times, whereas the last input at time $T$ is processed only once. This creates an inherent gap in the way that each time step influences the network weights during training. 

In this work, we purpose an alternative RNN framework. Instead of using a learned set of parameters to determine the transition between subsequent hidden states, our method uses a parameter-free shift mechanism, in order to distinguish between inputs fed at different times. This shift mechanism forms an orthogonal matrix operation, and is, therefore, not prone to the gradient explosion or to its vanishing. Furthermore, the various time steps are treated in a uniform manner, leading to an efficient and perhaps more balanced solution. This allows us, for example, to learn problems with much larger sequence lengths than reported in the literature. 

Our experiments show that our Shuffling Recurrent Neural Network (SRNN) is indeed able to tackle long-term memorization tasks successfully, and shows competitive results, in comparison to the current state of the art of multiple tasks. SRNN is elegant, easy to implement, efficient, and insensitive to its hyperparameters. We share our implementation at  \url{https://github.com/rotmanmi/SRNN}.

\section{Background and Related Work}
\label{sec:related}
RNNs have been the architecture of choice, when solving sequential tasks. The most basic structure, which we refer to as the vanilla RNN, updates at each time a hidden state vector $h_t$ using an input $x_t$,
\begin{equation}
h_t = \sigma\left( W_{1}h_{t-1} + W_{2}x_t \right) \equiv \sigma(z_t)\,,
\end{equation}
where $\sigma$ is a non-linear activation function, such as $\tanh$ or ReLU. Given a sequence of length $T$, $\left\{x_t\right\}_{t=1}^{T}$, computing the gradients of a loss function, $\mathcal{L}$, w.r.t to $W_1$,  $\frac{\mathcal{\partial L}}{\partial W_{1}}$, requires the application of the chain rule throughout all the hidden states $\left\{ h_t\right\}_{t=1}^T$,
\begin{equation}
\frac{\partial \mathcal{L}}{\partial W_{1}} = \sum_{t=1}^T{\frac{\partial \mathcal{L}}{\partial h_t} \frac{\partial h_t}{\partial W_{1}}} = \sum_{t=1}^T{\frac{\partial \mathcal{L}}{\partial h_t} \sigma'(z_t) h_{t-1}}\,,
\end{equation}
where  
$ \frac{\partial \mathcal{L}}{\partial h_t} = \sum_{t'=t+1}^{T}\frac{\partial \mathcal{L}}{\partial h_{t'}} \frac{\partial h_{t'}}{\partial h_t}$ and $ \frac{\partial h_{t'}}{\partial h_t}=   \prod_{i=t}^{t'} \frac{\partial h_{i+1}}{\partial h_i} = \prod_{i=t}^{t'} \sigma'(z_i)W_{1}$.
Depending on the maximal eigenvalue of $W_{1}$, the repeated multiplication by $W_{1}$ in $\frac{\partial \mathcal{L}}{\partial h_t}$ 
may lead to exponential growth or decay in the gradients of $\frac{\partial \mathcal{L}}{\partial W_{1}}$ when $T \gg 1$.

Many of the successful RNN methods utilize a gating mechanism, where the hidden state $h_t$ can be either  suppressed or scaled, depending on a function of the previous hidden state and the input. Among these solutions, there is the seminal Long-Short Term Memory network (LSTM) \cite{Hochreiter:1997:LSM:1246443.1246450} that utilizes a gating mechanism together with a memory cell, the powerful Gated Recurrent Unit (GRU) \cite{cho-etal-2014-learning}, and the recent Non-Saturating Recurrent Unit (NRU) \cite{chandar2019towards} that makes use of a non-saturating function, such as a ReLU, for the activation function. These units often make use of a gradient clipping scheme while training, since they contain no inherent mechanism to deal with exploding gradients over very long time sequences. 

A second family of RNNs focuses on constraining the weight matrix $W_1$ of the RNN to be orthogonal or unitary. Unitary Recurrent Neural Networks (uRNN)~\cite{urnn} force a strict structure regime on the parameter matrices, thus modifying these matrices in a sub-manifold of unitary matrices. Noting that the method neglects some types of unitary matrices, the Full-Capacity Unitary Recurrent Neural Network~\cite{fullurnn}, uses a weight parameterization that spans the complete set of unitary matrices, by constraining the gradient to reside on a Stiefel manifold. EUNN~\cite{jing2017tunable} employs a more efficient method in order to span this set. Another efficient approach to optimize in this space, which is based on the Householder reflections, was proposed in~\cite{mhammedi2017efficient}, and a Cayley transform parameterization was used in~\cite{helfrich2018orthogonal,maduranga2019complex}. The recent nnRNN~\cite{kerg2019non} method parameterizes the transformations between the successive hidden states using both a normal matrix and a non-normal one, where the first is responsible for learning long-scale dynamics and the second adds to the expressibility of the model.

\section{Method}
\label{sec:method}

\begin{figure}
\centering
\subfigure[]{\resizebox{0.40\linewidth}{!}{\begin{tikzpicture}[
    font=\sf \scriptsize,
    >=LaTeX,
    cell/.style={
        rectangle, 
        rounded corners=5mm, 
        draw,
        very thick,
        },
    operator/.style={
        circle,
        draw,
        inner sep=-0.5pt,
        minimum height =.4cm,
        },
    function/.style={
        ellipse,
        draw,
        inner sep=1pt
        },
    ct/.style={
        circle,
        draw,
        line width = .75pt,
        minimum width=1cm,
        inner sep=1pt,
        },
    gt/.style={
        rectangle,
        draw,
        minimum width=4mm,
        minimum height=3mm,
        inner sep=1pt
        },
    mylabel/.style={
        font=\scriptsize\sffamily
        },
    ArrowC1/.style={
        rounded corners=.25cm,
        thick,
        },
    ArrowC2/.style={
        rounded corners=.5cm,
        thick,
        },
    ]
     \node [cell, minimum height =4cm, minimum width=6cm] at (0,0){} ;
     
    \node[ct, label={[mylabel]below:Hidden}] (h) at (-4,-0) {$h_{t-1}$};
    \node[ct, label={[mylabel]left:Input}] (x) at (-0,-3) {$x_t$};
    
    \node[ct, label={[mylabel]below:Hidden}] (h2) at (4,0) {$h_{t}$};
    \node[ct, label={[mylabel]left:Output}] (x2) at (0,3) {$o_t = s\left(h_{t}\right)$};

    \node[operator, right=1.5cm of h, label={[mylabel]below:Circular shift}] (shift1) {$\gg$};
    
    \node[gt, above=1.0cm of x,minimum width=2cm] (nn) {$b$};
    \node[operator, above =of nn,right =of shift1] (sum1) {$+$};
    \node[operator, right=1.5cm of sum1,label={[mylabel]below:Non-linearity}] (act1) {$\sigma$};
    \draw [ArrowC1] (h) -- (shift1);
    \draw  [->, ArrowC2] (shift1) -- (sum1);
    \draw  [->, ArrowC2] (sum1) -- (act1);
    \draw  [->, ArrowC2] (nn.north) -|  (sum1.south);
    \draw [ArrowC1] (act1.east) -- (h2.west);
    \draw [ArrowC1] (act1.north) -| (x2.south);
    \draw [ArrowC1] (x) -- (nn);
    \end{tikzpicture}}\label{fig:SRNN}}
\hfill
\subfigure[]{\resizebox{0.40\linewidth}{!}{\begin{tikzpicture}[
		font=\sf \scriptsize,
		>=LaTeX,
		cell/.style={
			rectangle, 
			rounded corners=5mm, 
			draw,
			very thick,
		},
		operator/.style={
			circle,
			draw,
			inner sep=-0.5pt,
			minimum height =.4cm,
		},
		function/.style={
			ellipse,
			draw,
			inner sep=1pt
		},
		ct/.style={
			circle,
			draw,
			line width = .75pt,
			minimum width=1cm,
			inner sep=1pt,
		},
		gt/.style={
			rectangle,
			draw,
			minimum width=4mm,
			minimum height=3mm,
			inner sep=1pt
		},
		mylabel/.style={
			font=\scriptsize\sffamily
		},
		ArrowC1/.style={
			rounded corners=.25cm,
			thick,
		},
		ArrowC2/.style={
			rounded corners=.5cm,
			thick,
		},
		]
		\node[gt,minimum width=2cm,minimum height=1.5cm] at (0,0) (rep) {$f_r$};
		\node[ct, minimum width=0.5cm, text width=.399cm, align=flush center,line width = .25pt] at (3,0.5)(sigmoid) {{\tiny{sig\\moid}}};
		\node[gt, minimum width=2cm,minimum height=0.5cm] at (3,-0.5)(screen) {$W_s,b_s$};
		\node[gt, minimum width=1cm,minimum height=0.5cm] at (1.5,-1.5) (x) {$x_t$};
		\node[ct, minimum width=0.5cm] at (1.5,1) (piecewise) {$\odot$};
		\draw[-latex] (x) |- (1.5,-1) -|(screen);
		\draw[-latex] (x) |- (1.5,-1) -|(rep);
		\draw[-latex] (screen) -- (sigmoid);
		\draw[-latex] (rep) |-  (piecewise);
		\draw[-latex] (sigmoid) |- (piecewise);
		\draw[-latex] (piecewise) -- (1.5,1.55);
		\end{tikzpicture}} \label{fig:bstructure}}
\hfill
\caption{(a) The architecture of the SRNN layer. The output of $b$ is added to the shifted version of the previous hidden state, followed by a non-linearity, $\sigma$. The output at time $t$ is obtained by a function $s$ that is applied to the new hidden state. (b) The structure of network $b$. The primary sub-network $f_r$ is an MLP, the gating sub-network (right branch) has a single affine layer and a sigmoid non-linearity. The outputs of the two branches are multiplied elementwise to produce $b$'s output.}
\end{figure}
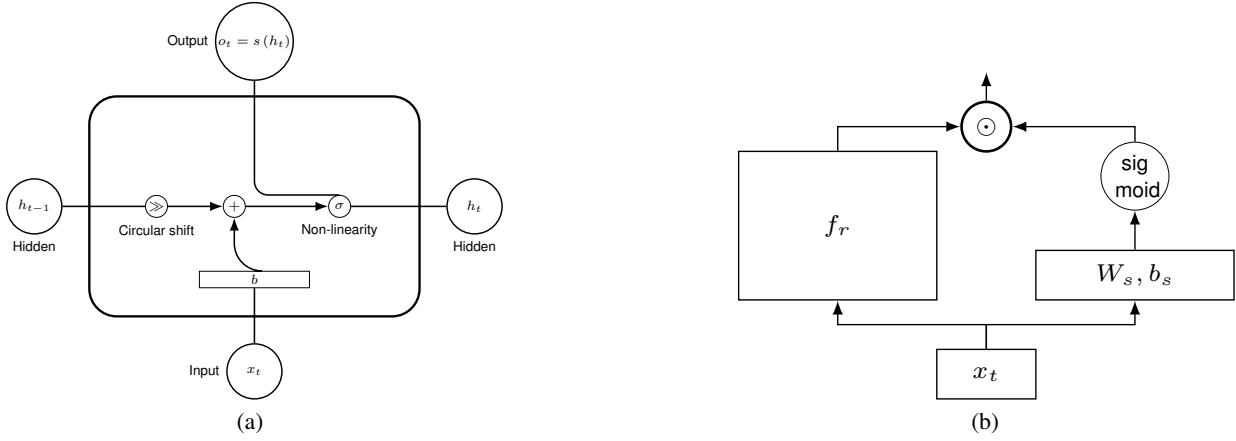

The SRNN layer contains two hidden-state processing components, the learned network $b$ that is comprised of fully connected layers, and a fixed permutation matrix $W_p$.  At each time step, the layer, like other RNNs, receives two input signals: the hidden state of the previous time step, $h_{t-1}\in\mathbb{R}^{d_h}$, and the input at the current time step, $x_t\in \mathbb{R}^{d_i}$, where $d_h$ and $d_i$ are the dimensions of the hidden state and the input, respectively. The following computation takes place in the layer (we redefine the notation, disregarding the definitions of Sec.~\ref{sec:related}):
\begin{equation}
\label{eq:srnn}
    h_{t} = \sigma \left( {{W_{p}}{h_{t-1}} + {b\left( {{x_t}} \right)}} \right) \equiv \sigma\left( z_t \right) \,,
\end{equation}
where $\sigma$ is the activation function (such as ReLU or $\tanh$), see Fig.~\ref{fig:SRNN}. The permutation operator $W_p$ reflects the time dependency, whereas $b$ is agnostic to the ordering of the input.  Without loss of generality, the permutation can be chosen as the shift operation that can represented by the off-diagonal matrix
\begin{equation}
    W_p = \scalemath{0.6}{{\left( {\begin{array}{*{5}{c}}
  0&1&0& \cdots &0 \\ 
  0& \ddots & \ddots & \ddots &0 \\ 
  \vdots& \ddots & \ddots & \ddots & \vdots  \\ 
  0& \ddots & \ddots & \ddots &1 \\ 
  1&0& \cdots &0&0 
\end{array}} \right)}}\,.
\end{equation}

Network $b$ employs a two-headed structure for a self-gating mechanism. The primary branch is composed of a MLP $f_r: \mathbb{R}^{d_i} \to \mathbb{R}^{d_h}$. The gating  branch scales the output of $b$ and contains a single affine layer with a sigmoid activation function.
\begin{equation}
b\left(x_t\right) = f_r\left( x_t \right) \odot \textnormal{sigmoid} \left( {W_s x_t + b_s}\right) \,,
\end{equation} 
where $W_s\in\mathbb{R}^{d_h\times d_i}$ and $b_s\in\mathbb{R}^{d_h}$ are the weights and biases of the gating branch. Fig.~\ref{fig:bstructure} depicts the structure of network $b$.

The output of the network at time step $t$, $o_t$, is obtained by using a single affine layer $s$,
\begin{equation}
	o_t = s\left( h_t \right) \,.
\end{equation}

{\bf Analysis of gradient dynamics\quad} Since $W_p$ is a permutation matrix, it is also orthogonal. Our method benefits from this property, since the successive applications of $W_p$ do not increase nor decrease the norm of the hidden state vector, such that the gradients with respect to the hidden state do not suffer from an exponential growth or decay. Therefore, the method can be applied without the use of any gradient clipping schemes. 

In addition, since the operator $W_p$ is not learned, and the weights of network $b$ do not appear with powers greater than one in the gradient. The gradient of the loss w.r.t a parameter $b_k$ in network $b$ is 
\begin{equation}
	\frac{\partial \mathcal{L}}{\partial {b_k}} = \sum_{t=1}^T \frac{\partial \mathcal{L}}{\partial {h_t}} \frac{\partial h_t}{\partial {b_k}} \,.
	\label{eq:lossdiv}
\end{equation}
The derivative of Eq.~\ref{eq:srnn} w.r.t $b_k$ yields a recursive equation,
\begin{equation}
	\frac{\partial h_t}{\partial {b_k}} = W_p\sigma'\left( z_t\right) \frac{\partial h_{t-1}}{\partial {b_k}} + \sigma'\left( z_t\right) \frac{\partial b\left( x_{t} \right)}{\partial b_k} \,.
	\label{eq:rec} 
\end{equation}
Expanding Eq.~\ref{eq:rec} yields the closed form,
\begin{equation}
	\frac{\partial h_t}{\partial {b_k}} = \sum_{i=1}^t \sigma'\left(z_i\right)\left(\prod_{j=1}^{i-1}\left[W_p\sigma'\left(z_j\right)  \right]\right)\frac{\partial b\left( x_i \right)}{\partial b_k}   \,.
	\label{eq:closed}
\end{equation}
Since $W_p$ is orthogonal, the only terms that may influence the gradients' growth or decay are $\sigma'\left(z_t\right)$. For most used activation functions, such as the sigmoid, $\tanh$ and ReLU, $\left\vert\sigma'\left(z_i\right) \right\vert$ is bounded by $1$. Therefore, one obtains 
\begin{equation}
	\left\vert \frac{\partial h_t}{\partial b_k} \right\vert \leq \left\vert\sum_{i=1}^t  \frac{\partial b\left( x_i \right)}{\partial b_k} \right\vert\,.
	\label{eq:grad}
\end{equation}

Combining this result with Eq.~\ref{eq:lossdiv} reveals that
\begin{equation}
    \left\vert \frac{\partial \mathcal{L}}{\partial {b_k}} \right\vert \leq \left\vert \sum_{t=1}^T \frac{\partial \mathcal{L}}{\partial {h_t}} \sum_{i=1}^t  \frac{\partial b\left( x_i \right)}{\partial b_k}  \right\vert \,.
\end{equation}
As $\frac{\partial \mathcal{L}}{\partial b_k}$ gains only linear contributions from the derivatives of $b$ it cannot explode.

For the ReLU activation function, the gradients in Eq.~\ref{eq:closed} vanish if and only if there is some time $t'$ in the future where $z_{t'} < 0$. Since previous hidden states can only increase $h_t$, negative contributions only arise due to the outputs of network $b$. These outputs, $b(x_t)$, approximately follow a normal distribution of $\mathcal{N}\left(0,  1\right)$~\cite{he2015delving}. An estimation for the number of time steps it takes for a neuron in $h_t$ to vanish is achieved by assuming that the probability of either obtaining a negative or a positive contribution to  its activation at step $t$ is $\frac{1}{2}$. This scenario is known as the Gambler's Ruin problem~\cite{steele2012stochastic}. Although the probability of having the hidden state vanish, $p\left(z_t \leq 0\right) = \frac{T}{1+T}$, approaches one for $T \gg 1$, the expectation value of the number of steps until this happens is $T$.

{\bf Time complexity\quad} The computation of $h_t$ does not involve any matrix multiplications between 
previous hidden-state $h_{t-1}$, and the permutation operator can be applied in  $O( d_h)$. The most time-consuming operator is the application of the function $b$. However, since $b$  does not require any information from previous states, it can be applied in parallel to all time steps, thus greatly reducing the total runtime. The time complexity, in comparison to the literature methods, is presented in Table~\ref{tab:timecomplexity} for a minibatch of size one. It assumes that the number of hidden units in each layer of $f_r$ is $O(d_i)$. As can be seen, our method is the only one that is linear in $d_h$.

\begin{table}[t]
    \begin{minipage}{.53\linewidth}
      \caption{Time complexity for one gradient step for sequence length $T$. $d_h$,$d_i$=hidden state and input dims.}
	\label{tab:timecomplexity}
	\begin{center}
		\begin{small}
			\begin{sc}
				\begin{tabular}{lc}
					\toprule
					Method & Time complexity\\
					\midrule
					Vanilla RNN     &$O\left(T d_h^2 + Td_hd_i  \right)$ \\
					GRU      & $O(T d_h^2+Td_hd_i)$ \\
					LSTM      & $O(T d_h^2+Td_hd_i)$ \\
					uRNN      &$O(T d_h \log d_h+Td_hd_i)$ \\
					NRU      &  $O(T d_h^2+Td_hd_i)$ \\
					SRNN    & $O(T d_h d_i)$  \\
					\bottomrule
				\end{tabular}
			\end{sc}
		\end{small}
	\end{center}
    \end{minipage}%
    \hfill
    \begin{minipage}{.45\linewidth}
      \centering
        \caption{Accuracy for the pMNIST and the bigger version, in which the MNIST image is embedded in a black image four times larger.}
\label{tab:psMNIST}
\begin{center}
\begin{small}
\begin{sc}
\begin{tabular}{lcc}
\toprule
Method &  pMNIST  & Big pMNIST \\
\midrule
LSTM     & 89.50\% & 33.60\% \\
GRU     & 91.87\% & 9.45\% \\
uRNN     & 91.74\% & Out of memory\\
NRU     & 91.64\% & 11.01\% \\
SRNN     & 96.43\% & 90.31\% \\
\bottomrule
\end{tabular}
\end{sc}
\end{small}
\end{center}
    \end{minipage} 
\end{table}

\section{Experiments}
\label{sec:experiments}
We compare our SRNN architecture to the leading RNN architectures from the current literature. The baseline methods include: (1) a vanilla RNN, (2) LSTM~\cite{Hochreiter:1997:LSM:1246443.1246450}, (3) GRU~\cite{cho-etal-2014-learning}, (4) uRNN~\cite{urnn}, (5) NRU~\cite{chandar2019towards}, and (6) nnRNN~\cite{kerg2019non}. All methods, except NRU, employed the RMSProp \cite{bengio2015rmsprop} optimizer with a learning rate of 0.001 and a decay rate of 0.9. For NRU, we have used the suggested ADAM ~\cite{kingma2014adam} optimizer with a learning rate of 0.001, and employed gradient clipping with a norm of one. The optimization parameters for nnRNN were taken from the official repository. For all problems involving one-hot inputs, we have added an embedding layer before the RNN, since it benefited all methods. We believe that reports, which have deemed GRU as ineffective in some of the proposed benchmarks, did not include such a layer. The activation function $\sigma$ in Eq.~\ref{eq:srnn} and within the network $b$ was a ReLU. The activation function of the vanilla RNN (denoted `RNN' in the figures) was $\tanh$.

{\bf Copying Memory Problem\quad}
RNNs are often challenged by the need to take into account information that has occurred in the distant past. The Copying Memory (MemCopy) task of~\cite{urnn} was designed to test the network's ability to recall information seen in the past. The objective is to memorize the first $10$ characters of the sequence. Let $A = \left\{a_i\right\}^8_{i=1}$ be a set of $8$ symbols, $a_9$ the blank symbol and $a_{10}$, the delimiter symbol. we create a sequence with the length $T+20$, where the first $10$ symbols are taken from $A$, then the next $T-1$ symbols are the blank symbol $a_9$ followed by a single appearance of the delimiter symbol, $a_{10}$. The last $10$ symbols are again set to the blank symbol $a_9$. The required output is a sequence of $T+10$ blank symbols followed by the first $10$ symbols taken from $A$ of the original sequence. The baseline model for this task predicts some constant sequence after the appearance of the delimiter symbol, $a_{10}$. The cross-entropy for this solution is $\frac{10 \ln 8}{T + 20}$. 


We trained all models with a minibatch of size $20$. We used a hidden size of $d_h=128$ for all models. For $f_r$ inside of the $b$ function of the SRNN we have used one hidden layer of size $8$, i.e., $f_r$ projects the input to activations in $\mathbb{R}^8$ and then to $\mathbb{R}^{128}$. 

\begin{figure}
	\begin{center}
		\includegraphics[width=0.95\linewidth]{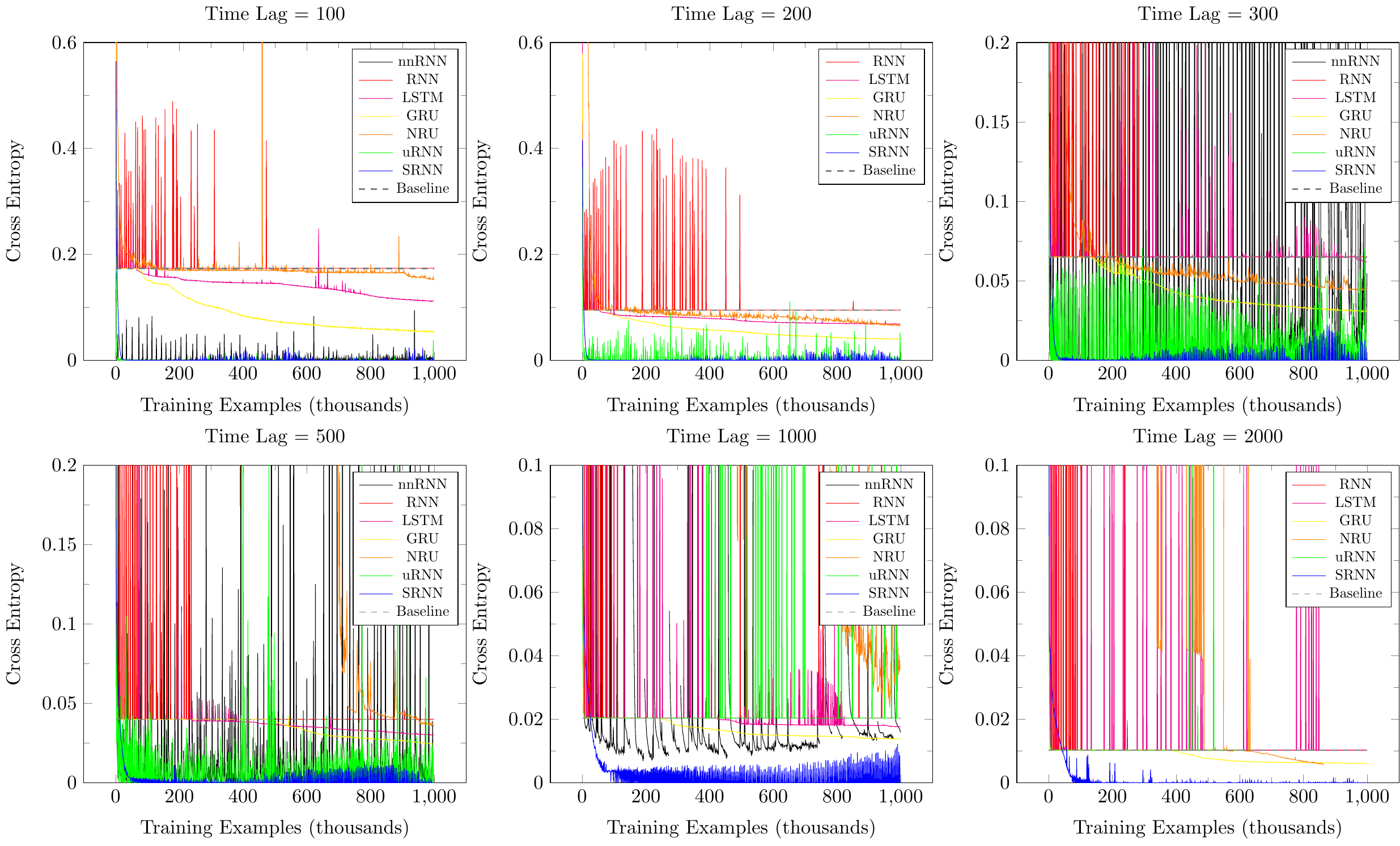}
	\end{center}
	\caption{Results for the MemCopy task. Shown is the cross entropy as a function of the number of training examples. Each plot depicted the results for a different sequence length. }
	\label{fig:copy}
\end{figure}

Fig.~\ref{fig:copy} shows the cross-entropy loss for all models for the time lag of $T=100,200,300,500,1000,2000$. While most methods are successful in dealing with sequences of length $T=100$, only the SRNN is able to deal efficiently with longer time lags. In contrast to results reported in the literature, the GRU and LSTM are able to partially solve this problem, and they even do so better than uRNN for sequences larger than 1000, where it starts to lose some of its stability. However, the convergence rate of LSTM and GRU is very slow. Note that the cross entropy obtained for SRNN, while better than other methods, can go up as training progresses. This is similar to what is observed in~\cite{urnn}, for their uRNN method, in the cases where uRNN is successful. nnRNN is successful for $T\leq 300$ and is unstable for longer sequences.

{\bf Adding Problem\quad}
The adding problem was first introduced in \cite{Hochreiter:1997:LSM:1246443.1246450}; We follow a close variant formulated in \cite{urnn}. The input for this task consists of two sequences of length $T$ (the two are concatenated to form $x_t$). The first sequence contains real numbers that have been uniformly sampled from $\mathcal{U}\left(0,1\right)$. The second sequence is an indicator sequence, that is set to $0$, except for two random entries that are set to $1$. One of these entries is located in the first half of the sequence and the other in the last half of the sequence. The objective of this task is to output the sum of the two numbers in the first sequence that correspond to the location of the $1$s in the second. The baseline to this task is the model which predicts 1, no matter what the input sequences are. The expected mean squared error (MSE) for this case is $0.167$. A hidden size of 128 was used for all methods. All models were fed with a minibatch of 50. As in the MemCopy problem, all the training samples were generated on the fly. Network $b$ of SRNN contains a hidden layer with size $8$.

\begin{figure}
	\begin{center}
		\includegraphics[width=0.95\linewidth]{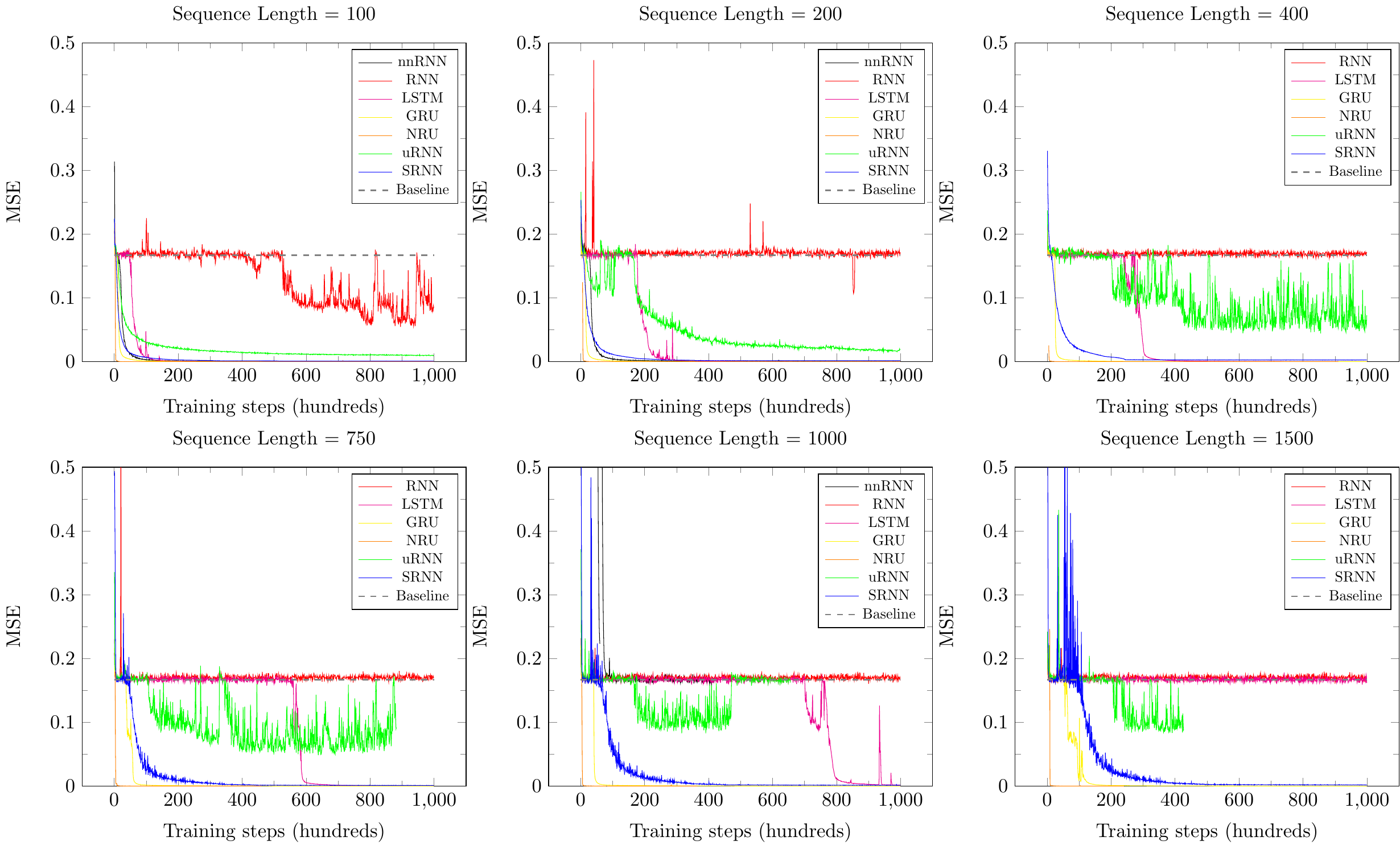}
	\end{center}
	\caption{Results for the adding problem for varying sequence lengths, noted on the top of each plot. Shown are MSE as a function of the number of training steps for our method and the baseline methods. For a longer sequence length, we sometimes stopped the run of uRNN before the allocated number of training iterations, due to the slow runtime of this method.}
	\label{fig:adding}
\end{figure}

Fig.~\ref{fig:adding} shows the MSE for all models for sequence lengths of $T=100,200,400,750,1000,1500$. NRU, GRU and SRNN solve this problem quickly, with NRU showing very fast convergence. LSTM is also successful in solving this task, but its convergence is much slower. nnRNN is able to solve this task whenever it was able to initialize properly for sequence sizes shorter than $200$.

{\bf Permuted Sequential MNIST\quad} The permuted MNIST (pMNIST) benchmark by \cite{le2015simple} measures the performance of RNNs, when modeling complex long-term dependencies. In this task, each MNIST image is reshaped into a 784-dimensional vector. This vector is fed to the RNN one entry at a time. The objective is to classify the image at the last time step. In order to increase the difficulty of the problem, a fixed permutation matrix is applied on each of the vectors before feeding them to the neural network. The same permutation and samples were used to train and evaluate all methods. 

In this task, all models had the order of 165k parameters, except for SRNN that had 50k parameters. We tried to initialize nnRNN with the same initialization presented in the literature of 800k parameters. However, it failed, so it was trained with less. Unfortunately, we were not able to get the vanilla RNN to converge on this task and nnRNN initialization (which depends on a numerical solver) failed, despite multiple efforts done using the official code of this method. 

For SRNN, a hidden size of $d_h = 1024$ was used, and the function $f_r$ of network $b$ contained three hidden layers of size $32$. A minibatch size of $100$ was used for training, similar to the experiments performed by NRU. Models were trained for $60$ epochs. Table~\ref{tab:psMNIST} presents the accuracy over the test set for the epoch with the lowest cross entropy over the validation set. As can be seen, SRNN is more effective than all other methods. We cannot compare with IndRNN~\cite{li2018independently}, which uses a different split.

In order to verify that our method is not overly sensitive to its parameters, namely the dimensionality of the hidden state and the depth and number of hidden units per layer of network $f_r$ (the primary sub-network of $b$), we tested multiple configurations. The results are reported in Fig.~\ref{fig:pmnistablation} and show that the method's accuracy is stable with respect to its parameters. As an ablation, we also report results in which gating is not used (the effect on the number of parameters is negligible). Without gating, performance somewhat decreases, especially for larger networks. However, overall, gating does not seem to be the most crucial part of the architecture.

\begin{figure}
\centering
\subfigure[]{\begin{tikzpicture}[scale=.75, trim axis left, trim axis right]
\begin{axis}[name=plot1,minor tick num=1,
xlabel={Number of hidden layers in $f_r$},ylabel={Accuracy},title={},restrict y to domain=0:10,ymin=0.94,ymax=0.97,xtick={1,2,3},ylabel shift= -10pt,legend style={nodes={scale=0.95, transform shape},at={(0.5,1.05)},
anchor=south,legend columns=2}]

\addlegendimage{empty legend}
 \addlegendentry{with gating}
 
 \addlegendimage{empty legend}
 \addlegendentry{w/o gating}
\addplot [red,line width=0.75pt,mark=x] coordinates {
	(1,0.9543)
	(2,0.9491)
	(3,0.9584)
	};
\addlegendentry{$d_h = 512$ , $\left\vert h_{f_r} \right\vert = 32$}

\addplot [red,line width=0.75pt,mark=x,dotted] coordinates {
	(1,0.9531)
	(2,0.9575)
	(3,0.9569)
	};
\addlegendentry{$d_h = 512$ ,$\left\vert h_{f_r} \right\vert =  32$}

\addplot [blue,line width=0.75pt,mark=x] coordinates {
	(1,0.9539)
	(2,0.9497)
	(3,0.9559)
};
\addlegendentry{$d_h = 512$, $f_r$ $\left\vert h_{f_r} \right\vert = 64$}

\addplot [blue,line width=0.75pt,mark=x,dotted] coordinates {
	(1,0.9478)
	(2,0.9542)
	(3,0.9550)
};
\addlegendentry{$d_h = 512$, $\left\vert h_{f_r} \right\vert = 64$}

\addplot [green,line width=0.75pt,mark=x] coordinates {
	(1,0.9594)
	(2,0.9638)
	(3,0.9643)
};
\addlegendentry{$d_h = 1024$, $\left\vert h_{f_r} \right\vert = 32$}

\addplot [green,line width=0.75pt,mark=x,dotted] coordinates {
	(1,0.9586)
	(2,0.9610)
	(3,0.9584)
};
\addlegendentry{$d_h = 1024$, $\left\vert h_{f_r} \right\vert = 32$}

\addplot [black,line width=0.75pt,mark=x] coordinates {
	(1,0.9579)
	(2,0.9598)
	(3,0.9603)
};
\addlegendentry{$d_h = 1024$, $f_r$ $\left\vert h_{f_r} \right\vert = 64$}


\addplot [black,line width=0.75pt,mark=x,dotted] coordinates {
	(1,0.9592)
	(2,0.9599)
	(3,0.9585)
};
\addlegendentry{$d_h = 1024$, $\left\vert h_{f_r} \right\vert = 64$}

\end{axis}
\end{tikzpicture}
\label{fig:pmnistablation}}
\subfigure[]{\begin{tikzpicture}[scale=.75, trim axis right]
\begin{axis}[name=plot2,minor tick num=1,
xlabel={Number of hidden layers in $f_r$},ylabel={MSE},title={},xtick={1,2,3},legend style={nodes={scale=0.95, transform shape},at={(0.5,1.05)},
anchor=south,legend columns=2},ymin=0,ymax=25]
\addplot [red,line width=0.75pt,mark=x] coordinates {
	(1,18.29)
	(2,16.45)
	(3,22.54)
	};
\addlegendentry{$d_h = 128$ , $\left\vert h_{f_r} \right\vert =  8$}

\addplot [red,dashed,line width=0.75pt,mark=x] coordinates {
	(1,11.82)
	(2,12.02)
	(3,12.69)
};
\addlegendentry{$d_h = 128$ , $\left\vert h_{f_r} \right\vert =  32$}

\addplot [blue,line width=0.75pt,mark=x] coordinates {
	(1,9.32)
	(2,10.16)
	(3,11.26)
};
\addlegendentry{$d_h = 256$ , $\left\vert h_{f_r} \right\vert =  8$}

\addplot [blue,dashed,line width=0.75pt,mark=x] coordinates {
	(1,7.81)
	(2,7.95)
	(3,7.85)
};
\addlegendentry{$d_h = 256$ , $\left\vert h_{f_r} \right\vert = 32$}

\addplot [green,line width=0.75pt,mark=x] coordinates {
	(1,4.64)
	(2,3.72)
	(3,6.41)
};
\addlegendentry{$d_h = 512$ , $\left\vert h_{f_r} \right\vert =  8$}

\addplot [green,dashed,line width=0.75pt,mark=x] coordinates {
	(1,3.64)
	(2,3.72)
	(3,4.73)
};
\addlegendentry{$d_h = 512$ , $\left\vert h_{f_r} \right\vert = 32$}

\addplot [black,line width=0.75pt,mark=x] coordinates {
	(1,1.69)
	(2,1.45)
	(3,2.11)
};
\addlegendentry{$d_h = 1024$ , $\left\vert h_{f_r} \right\vert =  8$}

\addplot [black,dashed,line width=0.75pt,mark=x] coordinates {
	(1,1.52)
	(2,0.88)
	(3,1.36)
};
\addlegendentry{$d_h = 1024$ , $\left\vert h_{f_r} \right\vert = 32$}

\addplot [purple,line width=0.75pt,mark=x] coordinates {
	(1,1.52) 
	(2,1.45)
	(3,1.33)
};
\addlegendentry{$d_h = 2048$ , $\left\vert h_{f_r} \right\vert = 8$}

\addplot [purple,dashed,line width=0.75pt,mark=x] coordinates {
	(1,0.71)
	(2,0.64)
	(3,0.73)
};
\addlegendentry{$d_h = 2048$ , $\left\vert h_{f_r} \right\vert = 32$}
\end{axis}
\end{tikzpicture}
\label{fig:timitablation}}
\caption{(a) SRNN accuracy on pMNIST for multiple configurations. For each configuration, we report the accuracy for up to three hidden layers. Dashed lines employ no gating in network $b$. (b) Same for TIMIT, reporting MSE, for varying $d_h$ or the hidden state of $f_r$ in $b$.}
\end{figure}
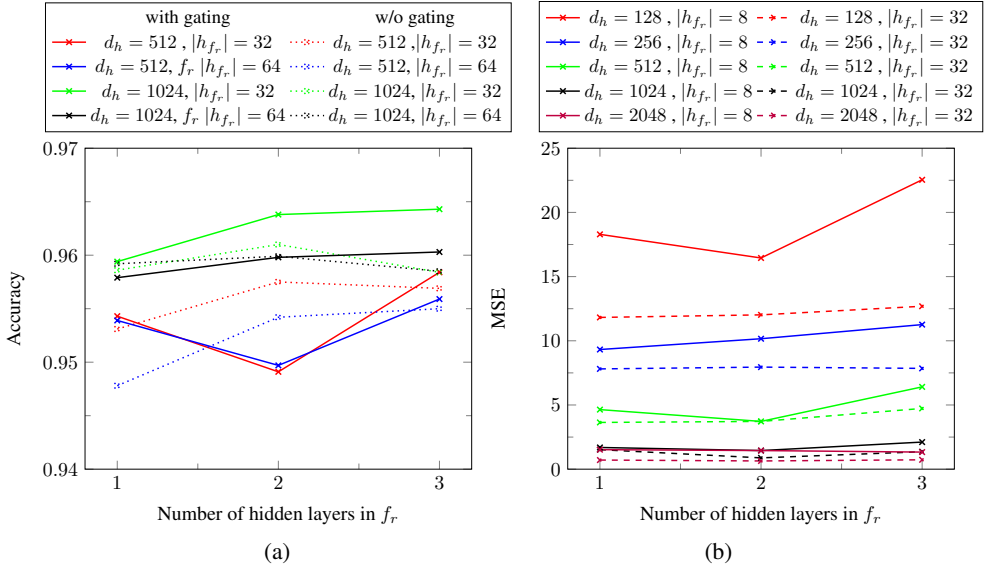
\begin{table}[t]
\caption{MSE on the TIMIT for various algorithms. $d_h$ is the size of the hidden state. Src=source.}
\label{tab:timit}
\smallskip
\begin{small}
\begin{sc}
\begin{tabular}{|c@{~}lcc|c@{~}lcc|c@{~}lcc|}
\toprule
Src & Method &  $d_h$ & MSE&Src & Method &  $d_h$ & MSE&Src & Method &  $d_h$ & MSE\\
\midrule
{\multirow{12}{*}{\cite{pmlr-v97-lezcano-casado19a}}} &  LSTM & 84& 14.30 & {\multirow{4}{*}{\rotatebox[origin=c]{0}{\cite{pmlr-v97-lezcano-casado19a}}}} & RGD &128 &14.58& \parbox[t]{2mm}{\multirow{12}{*}{\rotatebox[origin=c]{90}{Our runs}}} &uRNN     & 128 & 12.09\\
& LSTM & 120&12.95 & & RGD &128 &14.58& &uRNN     & 256 & 10.83\\
 & LSTM & 158&12.62 & & RGD &192 &14.50& &uRNN     & 512 & 11.90\\
\global\setlength\aboverulesep{0pt}
\global\setlength\belowrulesep{0pt}
& EXPRNN & 224& 5.30 & & RGD &256 &14.69& &NRU     & 128 &  12.26\\
    \cmidrule(l){5-8}
 & EXPRNN& 322 &4.38 & \parbox[t]{2mm}{\multirow{8}{*}{\rotatebox[origin=c]{90}{Our runs}}}
 &LSTM     & 128 & 19.78& &NRU     & 256 & 5.90 \\
\global\setlength\aboverulesep{0.4ex}
\global\setlength\belowrulesep{0.65ex}

 & EXPRNN& 425& 5.48& &LSTM     & 256 & 15.91& &NRU     & 512 & 3.23 \\
 & SCORNN& 224&8.50& &LSTM     & 512 & 12.32& &NRU     & 1024 & 0.35\\
 & SCORNN& 322&7.82& &LSTM     & 1024 & 8.66& &SRNN     & 128 & 12.69 \\
 & SCORNN& 425&7.36& &GRU     & 128 & 39.66  & &SRNN     & 256 & 7.85  \\
 & EURNN&158 &18.51& &GRU     & 256 & 37.17 & &SRNN     & 512 & 4.73  \\
& EURNN&256 &15.31& &GRU     & 512 & 33.00 & &SRNN     & 1024 & 1.34\\
 & EURNN &378 &15.15& &GRU     & 1024 & 26.53& &SRNN     & 2048 & 0.64\\
 \bottomrule
\end{tabular}
\end{sc}
\end{small}
\end{table}

Another version of pMNIST was tested, in which the MNIST image is padded with zeros, so that it is four times as large. The results are also reported in Table~\ref{tab:psMNIST}. Evidently, our method has an advantage in this dataset, which requires a much larger amount of memorization. Note that uRNN could not run due to GPU memory limitations, despite an effort to optimize the published code using TorchScript (\url{https://pytorch.org/docs/stable/jit.html}).

{\bf TIMIT\quad}
The TIMIT~\cite{garofolo1993darpa} speech frames prediction task was introduced by \cite{fullurnn} and later (following a fix to the way the error is computed) used in~\cite{pmlr-v97-lezcano-casado19a}. The train / validation/ test splits and exactly the same data used by previous work are employed here: using 3640 utterances for the training set, 192 for the validation set, and a test set of size 400. 

Table~\ref{tab:timit} reports the results of various methods, while varying the dimensionality of the hidden state, both for our runs, and for baselines obtained from~\cite{pmlr-v97-lezcano-casado19a}. Despite some effort, we were not able to have the published code of nnRNN run on this dataset (failed initialization). Running uRNN for dimensionality larger than 512 was not feasible. 
As can be seen, our SRNN outperforms almost all of the baseline methods, including EXPRNN~\cite{pmlr-v97-lezcano-casado19a}, SCORNN~\cite{helfrich2018orthogonal}, EURNN~\cite{jing2017tunable}, and RGD~\cite{fullurnn}. The only method that outperforms SRNN is NRU. However, the NRU architecture has an order of magnitude more parameters than our method for a given hidden state size; NRU with $d_h=1024$ has more than ten times the number of parameters than SRNN with $d_h=2048$. 

The SRNN model used on TIMIT has three hidden layers in $f_r$, each with 32 hidden neurons in each. However, the method is largely insensitive to these parameters, as can be seen in Fig.~\ref{fig:timitablation}. Networks with 8 hidden neurons in each layer perform slightly worse than those with 32, and the number of layers has a small effect when the dimensionality of the hidden state is high enough.

\begin{table}[t]
\caption{The runtime in seconds for one training epoch for various methods on a minibatch of 100. For MemCopy and Adding $T=300$, SRNN's $b$ network contains one hidden layer in $f_r$ with $32$ units. The number of samples in MemCopy epoch is $1000$, and the number of samples in Adding epoch is $10000$. In all cases $d_h=128$. The same machine was used for all experiments. }
\smallskip
\label{tab:runtime}
\centering
\begin{small}
\begin{sc}
\begin{tabular}{|@{~}l@{~}c@{~}c@{~}c@{~}c@{~~}c|l@{~}c@{~}c@{~}c@{~}c@{~~}c@{~}|}
\toprule
\global\setlength\aboverulesep{0pt}
\global\setlength\belowrulesep{0pt}
& \multicolumn{4}{c}{Runtime per dataset (sec)} & param& & \multicolumn{4}{c}{Runtime per dataset (sec)} & param\\
\cmidrule{2-5}
\cmidrule{8-11}
\global\setlength\aboverulesep{0.4ex}
\global\setlength\belowrulesep{0.65ex}
 &  MemCpy & Add & pMNIST & TIMIT & Add& &  MemCpy & Add & pMNIST & TIMIT & Add\\
\midrule
RNN & 0.18& 1.53& 19.64 & 1.76& 17k & NRU &    7.13 & 61.77& 835.47&  42.98 &102k\\
					LSTM  & 0.26  & 1.76 & 22.88&  1.80 & 67k& uRNN  &  13.48 & 112.78 & 1510.54& 76.10 & {\bf 2k}\\
					GRU   &  0.19 & 1.63 & 17.43& 1.78  & 50k& SRNN  &  {\bf 0.07}& {\bf 0.77}& {\bf 8.59}&  {\bf 1.27} &5k \\
\bottomrule
\end{tabular}
\end{sc}
\end{small}
\vspace{-.4cm}
\end{table}

\begin{figure}[t]
    \centering
    \begin{tabular}{cc}
    	\def\xaddd{2.395 * 0.40}
        \def\yaddd{-4.875 * 0.40}
    	\begin{tikzpicture}
	\node[label={west:$h_t$},inner sep=0pt,label={north:$t\rightarrow$}] (hidden)  at (0,0) {\includegraphics[width=0.4\textwidth]{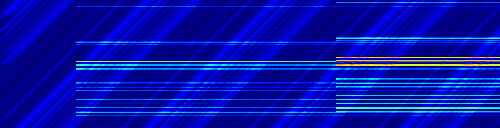}};
	\draw[-latex] (\xaddd,-1.2) -- node[right=5pt]{$t=336$} (\xaddd,-0.81);
	\draw[-latex] (\yaddd,-1.2) -- node[right=5pt]{$t=76$}(\yaddd,-0.81);
	\end{tikzpicture} & \multirow[vpos]{3}{*}[0.805in]{
 	\begin{tikzpicture}
 	\node[label={west:$h_t$},label={north:$t\rightarrow$},inner sep=0pt] (hidden)  {\includegraphics[width=0.46\textwidth]{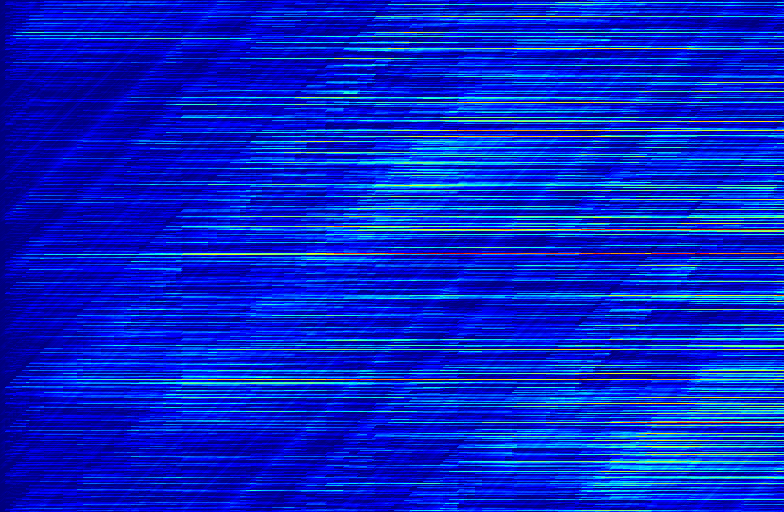}};
 	\end{tikzpicture}} \\
	\quad\quad{\small(a)} & \\
	\,\,\,\begin{tikzpicture}
	\node[label={west:$h_t$},inner sep=0pt,label={north:$t\rightarrow$}] (hidden)  at (0,0) {\includegraphics[width=0.4\textwidth]{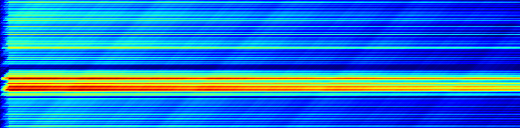}};
	\end{tikzpicture}&\\
	\quad\quad{\small(b)} &\quad\quad{\small(c)} \\
	\end{tabular}
    \caption{(a) The shifted hidden state of the SRNN for the Adding task.  The arrows indicate the positions where the indicator sequence in $x_t$ is set to $1$. These positions split the hidden state into three regions. (b) The shifted hidden state of the SRNN for the MemCopy task. After $T=10$ there is almost no change to the hidden state, as expected. (c) The shifted hidden state of the SRNN for the pMNIST task. Unlike the Adding or the MemCopy Task, the hidden state evolves through time, and the cell is able to suppress activations or accumulate additional information.}
     \label{fig:pmnisthiddenstate}
\end{figure}

{\bf Visualization of the hidden units\quad}
Fig.~\ref{fig:pmnisthiddenstate} provides a visualization of the  hidden state, $h_t$ of the SRNN for the Adding problem and the pMNIST task. For better visualization, the plot has been modified so the shift is removed. As can be seen, the method can retain information in cases in which memorization is needed, such as in the Adding problem. It also has the capacity to perform non-trivial computations in cases in which processing is needed, such as the pMNIST. SRNN also does not suffer from the vanishing gradient problem, since the activations are maintained over time. 

\begin{figure}[t]
    \centering
    \includegraphics[width=0.1\linewidth]{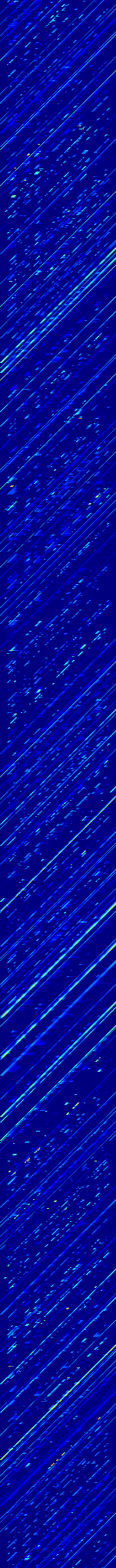}
    \caption{The shifted hidden state of the SRNN with $2048$ units for the TIMIT task with a sequence length of $T=151$. }
    \label{fig:timithiddenstate}
\end{figure}

Fig.~\ref{fig:timithiddenstate} provides a visualization of the hidden state, $h_t$ of the SRNN for TIMIT. In this example $h_t$ contains $2048$ units.
SRNN is able to remove irrelevant information for the future from the hidden state, an ability that is not required for the addition or the Memory Copy Tasks.

\smallskip
{\bf Capacity\quad}
Since the orthogonal matrix SRNN employs is fixed, one may wonder if it is less expressive than other RNNs, or if it has less capacity. The experiments below indicate that the capacity of SRNN is comparable to that of other methods with the same number of hidden units (although by avoiding the usage of complex cells, the number of parameters is smaller).

One way to measure the SRNN capacity is to measure its ability to differentiate between samples taken from the same probability distribution. In order to create an artificial differentiation, all the labels of MNIST are shuffled, and then a subset of $N$ samples is randomly selected. Since the focus of this experiment is not assessing  the ability of the SRNN or other baseline models to tackle long sequences, the MNIST images are cropped to a $8\times8$ or $16\times16$ center patches, so the generated sequences are now $64$ or $256$ steps long. For this experiment the models were selected to have an order of $15K$ parameters.  Testing is done on the training samples, since the aim is to measure the ability overfit random labels.

\begin{figure}[t]
    \centering
    \includegraphics[width=\linewidth]{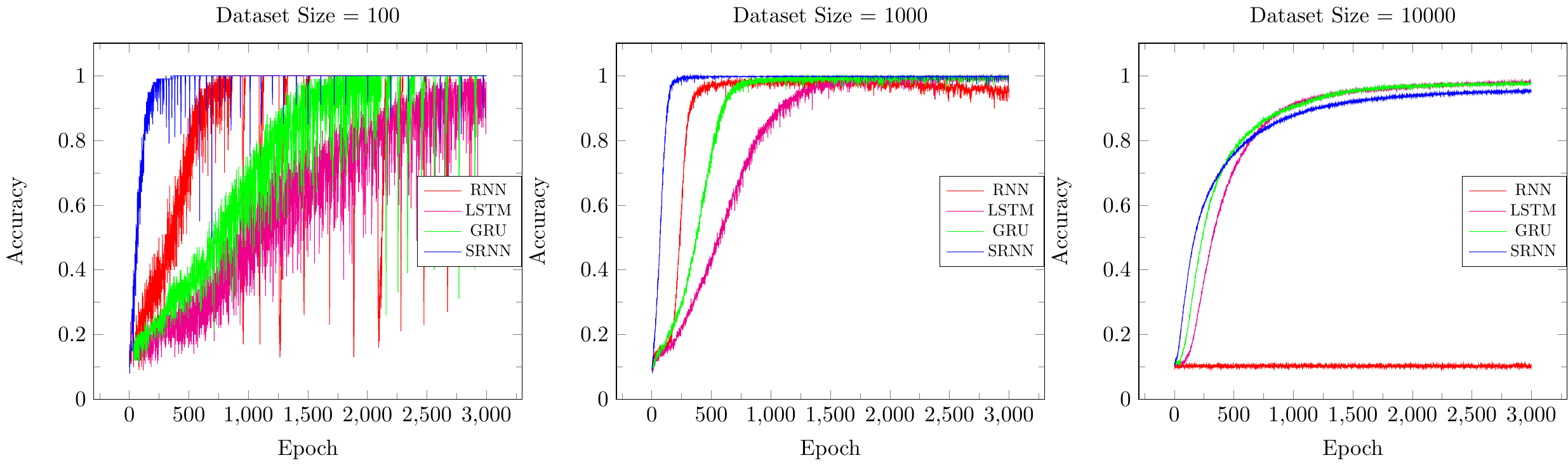}
    \caption{Fitting random labels for $N=100,1000,10000$ samples of $8\times8$ cropped from MNIST images.}
    \label{fig:express}
\end{figure}

Fig.~\ref{fig:express} presents the classification accuracy with respect to the random label for the SRNN, LSTM, GRU and the vanilla RNN architectures for different subset sizes,  $N=100,1000,10000$ and cropped patches of size $8\times8$. As can be seen, for $N=100,1000$, all models are able to differentiate between the different samples, i.e., overfit to the random labels. For the case of $N=10000$, all models succeed, except for the vanilla RNN. 

\begin{figure}[t]
    \centering
    \includegraphics[width=\linewidth]{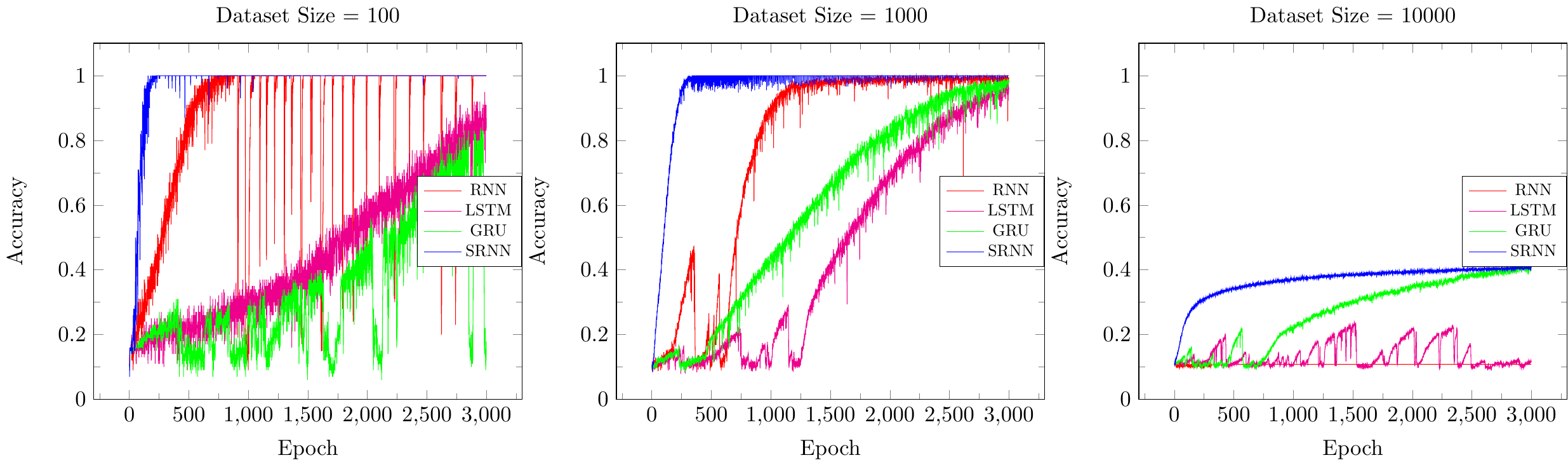}
    \caption{Fitting random labels for $N=100,1000,10000$ samples of $16\times16$ cropped from  MNIST images.}
    \label{fig:expressv3}
\end{figure}

Fig.~\ref{fig:expressv3} depicts a similar experiment for the case of $16\times16$ patches. For $N=100$ and $N=1000$ all methods are able to overfit, however, SRNN shows a marked advantage in the number of required training epochs. For $N=10,000$, only SRNN and GRU are able (partially) overfit the data. 

In addition, it is evident (in both settings) that as the sequence length increases, SRNN fits the data faster, which suggests that it has a high capacity and/or that it trains more effectively.

\smallskip
{\bf Runtime\quad} Table~\ref{tab:runtime} compares the runtime of our method, for a single epoch on each listed dataset, in comparison to the official implementations of other methods. As can be seen, our method is much more efficient than the other methods. In the same set of experiments, we also recorded the number of parameters reported by pytorch. This number is the same across benchmarks, except for the embedding size and the results reported are from the adding problem, where the embedding size is negligible. The only method with fewer parameters than our method is uRNN, which parameterizes the unitary transformation very efficiently.

\section{Conclusions}
\label{sec:conclusions}
In this work, we introduce a new RNN architecture that does not suffer from vanishing and exploding gradients. The method employs gating only on the sub-network that processes the input, makes use of a fixed shifting operator as an orthogonal transformation of the hidden states, is efficient, and has a lower complexity than other RNN architectures. The new method obtains competitive results in comparison with previous methods, which rely on much more sophisticated machinery.

\section*{Acknowledgments}
Tel Aviv University (TAU) has received funding for this project from the European Research Council (ERC) under the European Unions Horizon 2020 research and innovation program (grant ERC CoG 725974). The authors would like to thank Amit Dekel for valuable insights and Daniel Dubinsky for the CUDA-optimized implementation. The contribution of the first author is part of a Ph.D. thesis research conducted at TAU.

\bibliography{rnn}
\bibliographystyle{plain}

\end{document}